\pdfoutput=1
\documentclass[letterpaper]{article} 
\usepackage[]{aaai25}  
\usepackage{times}  
\usepackage{helvet}  
\usepackage{courier}  
\usepackage[hyphens]{url}  
\usepackage{graphicx} 
\usepackage{natbib}  
\usepackage{caption} 
\graphicspath{{figs/}}

\usepackage{amsmath}
\usepackage{amssymb}
\usepackage{algorithm}
\usepackage{algpseudocode}

\usepackage[frozencache,cachedir=.]{minted}

\usepackage{spverbatim}

\usepackage{oplotsymbl}

\usepackage{newfloat}
\usepackage{listings}
\DeclareCaptionStyle{ruled}{labelfont=normalfont,labelsep=colon,strut=off} 
\floatstyle{ruled}
\newfloat{listing}{tb}{lst}{}
\floatname{listing}{Listing}
\title{Inductive Learning of Logical Theories with LLMs:\\ An Expressivity-Graded Analysis}
\author {
    João Pedro Gandarela\textsuperscript{\rm 1},
    Danilo Carvalho\textsuperscript{\rm 2},
    André Freitas\textsuperscript{\rm 1,2,3}
}
\affiliations {
    \textsuperscript{\rm 1}Idiap Research Institute\\
    \textsuperscript{\rm 2}National Biomarker Centre, CRUK-MI, University of Manchester\\
    \textsuperscript{\rm 3}Department of Computer Science, University of Manchester\\
    \{firstname.lastname\}@idiap.ch, \{firstname.lastname\}@manchester.ac.uk
}

\begin{document}

\maketitle

\begin{abstract}
This work presents a novel systematic methodology to analyse the capabilities and limitations of Large Language Models (LLMs) with feedback from a formal inference engine, on logic theory induction. The analysis is complexity-graded w.r.t. rule dependency structure, allowing quantification of specific inference challenges on LLM performance.
Integrating LLMs with formal methods is a promising frontier in the Natural Language Processing field, as an important avenue for improving model inference control and explainability. In particular, inductive learning over complex sets of facts and rules, poses unique challenges for current autoregressive models, as they lack explicit symbolic grounding. While they can be complemented by formal systems, the properties delivered by LLMs regarding inductive learning, are not well understood and quantified.
Empirical results indicate that the largest LLMs can achieve competitive results against a SOTA Inductive Logic Programming (ILP) system baseline, but also that tracking long predicate relationship chains is a more difficult obstacle than theory complexity for LLMs.
\end{abstract}

%

\section{Introduction}

The integration of Large Language Models (LLMs) with formal methods stands out as a promising frontier in the field of Natural Language Processing. It is an avenue for improving model inference control and explainability, by both complementing the content flexibility of Large Language Models (LLMs) with the systematicity of symbolic/formal systems \cite{quan2024enhancing,quan2024verification} and by using well-defined formal settings to assess the underlying inference properties of the model.

Inductive Logic Programming (ILP) is a subfield of symbolic AI which focuses on methods that can derive (generalise) theories to explain observed facts and rules \cite{ilpmugg, ilpwolf}. Addressing inductive learning over complex sets of facts and rules, poses unique challenges for current autoregressive LLMs, as they do not operate over data symbolically, rather combining an extensive set of structural an semantic signals to approximate the most probable answer in a generative fashion. 

While such means of problem solving might lack explicit symbolic grounding, LLMs can leverage its large-scale internal representation to support inductive-style inference. Still, the properties delivered by LLMs w.r.t. inductive learning, in particular regarding logic rules and theory induction, are not well understood and quantified.

This paper presents a systematic methodology to evaluate the inductive learning properties (in the context of logic theory induction) of LLMs. It is aimed at answering the following research questions (RQs):

\noindent \textbf{RQ1.} To what extent the combination of an LLM with feedback from a formal inference engine can compare to a SOTA inductive logic programming (ILP) system in logic theory induction, w.r.t. inference quality at different expressivity levels?

\noindent \textbf{RQ2.} How does the expressivity of the target theories affect inference quality of LLMs for inductive reasoning?\\

\begin{figure*}[ht!]
    \centering
    \includegraphics[width=\linewidth]{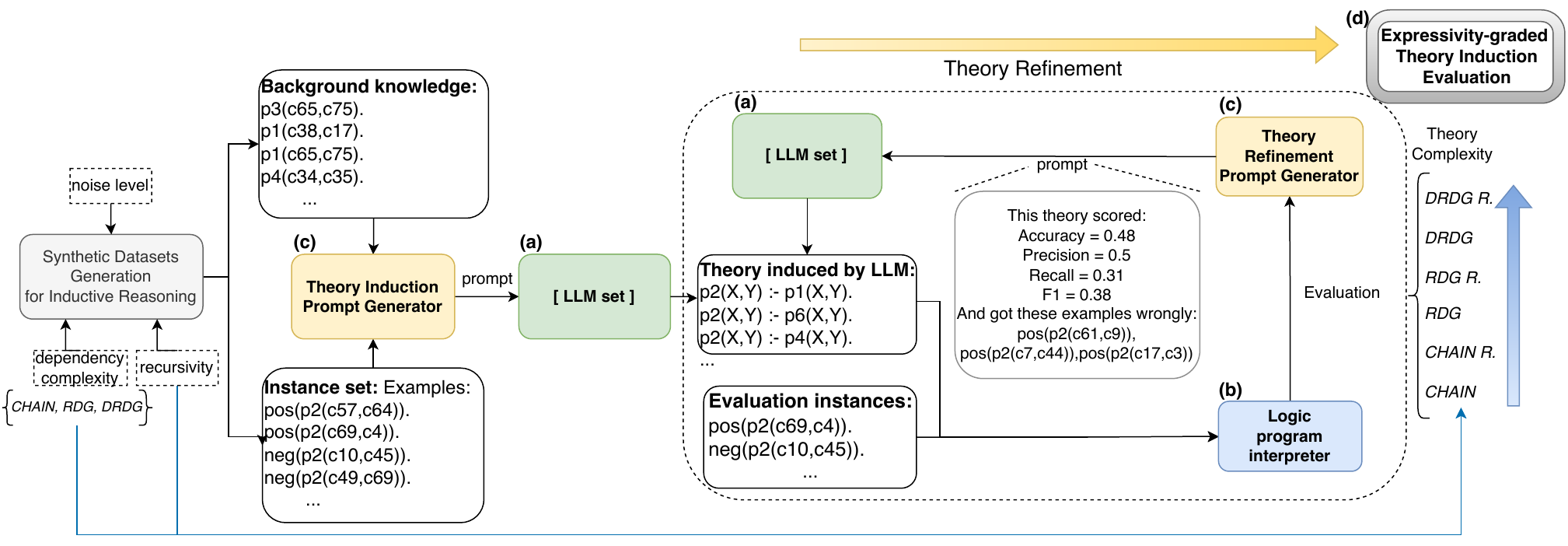}
    \caption{The proposed method to evaluate theory induction with an LLM in Prolog based on background knowledge and training examples. The process starts with a prompt generator $(c)$ that formulates prompts for an LLM $(a)$. Both the background knowledge and training sets are parameterised by different noise and rule expressive power levels: Chain, Rooted Directed Graph (DG), Disjunctive Rooted DG, and Mixed. The LLM generates theories, which are then evaluated by a logic program interpreter $(b)$. The evaluation feedback, including accuracy, precision, recall, and F1 scores, as well as wrongly classified examples, is used to refine the prompts iteratively. We analyse and categorise the generated theories according to their expressive power $(d)$.}
    \label{fig:method}
\end{figure*}

In order to address these RQs, we propose a method for combining iterative theory refinement on stock LLMs (i.e., zero-shot inference~\cite{10.5555/3495724.3495883,radford2019language}), a formal ILP inference engine and a synthetic generator for inductive reasoning datasets, in order to perform systematic evaluations of the LLM induced theories, using state-of-the-art ILP solvers as a baseline. Moreover, to quantify the extent in which LLMs can address ILP tasks, the evaluation is graded w.r.t. the expressivity of the target rulesets. Figure~\ref{fig:method} schematises the proposed approach.

This work's main contributions are as follows:

\begin{enumerate}
    \item (Methodological) A novel method for systematic evaluation of LLM induced logic theories with feedback from a formal inference engine.
    \item (Empirical) A detailed empirical analysis on the strengths and limitations of SOTA LLMs regarding logic theory induction generation, according to target ruleset expressivity.
    \item (Resources) A reusable and extensible framework for extending and assessing the inductive capabilities of LLMs.
\end{enumerate}

The remainder of the paper is organised as follows: Section ``\textit{Inductive Learning, Expressive Power \& Datasets}'' formalises the relevant ILP concepts, dataset generation, and LLM used in the paper. Section ``\textit{Proposed Approach}'' describes the proposed method and algorithms. Section ``\textit{Experiments}'' presents the experimental procedures, results and discussion, followed by related work and conclusions.

\section{Inductive Learning, Expressive Power \& Datasets} \label{sec:bg}

In this section we introduce the target task and the typology of inductive learning expressivity levels, as well as the dataset generation process.

\subsection{Inductive Logic Programming}
Inductive Logic Programming main objective is to generate logical hypotheses or theories from the available background knowledge, such as facts, rules, and positive and negative examples. Unlike traditional machine learning methods, ILP works directly with logical predicates and rules, using a formal representation that is usually represented using first-order logic (FOL). For example, the fact ``parent(john, tom).'' means that John is the parent of Tom, and ``parent(john, anne).'' means that John is the parent of Anne. From that, we can create a rule (\textit{theory}) ``sibling(X, Y) :- parent(P, X), parent(P, Y), X $\ne$ Y.'' . This rule states that if there exists a parent who has both X and Y as children, and X and Y are not identical, then X and Y are considered siblings. Deriving rules from a set of examples is a process known as \textit{theory induction}. This task can be formally defined as follows:\\ 
\noindent
\textbf{Given:}

\begin{itemize}
    \item \textit{Background Knowledge (BK):} A set of logical clauses that represent prior knowledge about the domain.
    \item \textit{Positive Examples ($E^+$):} A set of ground facts (instances) which the learned theory should entail (i.e., these are examples that should be true according to the theory).
    \item \textit{Negative Examples ($E^-$):} A set of ground facts which the learned theory should not entail (i.e., these are examples that should be false according to the theory).
\end{itemize}

\noindent
\textbf{Goal:} Find a hypothesis \( H \) (a set of logical clauses) such that:

\begin{enumerate}
    \item \textbf{Completeness:} For every example \( e \in E^+ \):
    \[
    H \cup BK \models e
    \]
    Meaning the hypothesis \( H \), together with the background knowledge \( BK \), should entail all positive examples.

    \item \textbf{Consistency:} For every example \( e \in E^- \):
    \[
    H \cup BK \not\models e
    \]
    Meaning the hypothesis \( H \), together with the background knowledge \( BK \), should not entail any negative examples.
\end{enumerate}

\noindent
\textbf{Formally:} An ILP system seeks a hypothesis \( H \) that satisfies:
\[
\forall e \in E^+, \ BK \cup H \models e
\]
\[
\forall e \in E^-, \ BK \cup H \not\models e
\]

\noindent
\textbf{Conclusion:} The learned hypothesis \( H \) should thus be a logical theory that explains the positive examples and excludes the negative examples, based on the given background knowledge.






\subsection{Theory Expressivity}

Inductive learning can be organised according to different classes of expressive power, which involves a typology of the structural complexity of the problem. Moreover, variables such as the amount and type of noise within the evidence set (such as the number of incorrect or missing facts or completeness levels) can be integrated within this setting. Following the typology of \cite{cornelio_thost_rudas} four base categories of rules can be introduced based on their dependencies: Chain, Rooted Directed Graph (RDG), Disjunctive Rooted DG, and Mixed. Each category represents a hierarchical generalisation, with each step encompassing a broader range of structures. Starting with CHAIN, which represents a linear composition of predicates, RDG is a generalisation of CHAIN, meaning it includes all chain structures but also accommodates more complex dependencies. Moving up, DRDG generalises RDG by incorporating directed relationships, thus offering a more extensive representation of dependencies. Finally, Mixed is a DRDG with recursion, containing connected components from CHAIN, RDG, and DRDG. Each progression from CHAIN to MIXED represents a step towards greater inclusivity and complexity in the types of structures captured.

The expressive power of each class and their characteristics are summarised in Table~\ref{tab:complexity_cls}. A detailed description of each class is provided in the supplementary material\footnote{A version of this paper with the supplementary material can be found at \url{https://arxiv.org/pdf/2408.16779}. \label{fn:arxiv_appendix}}.

\begin{table}[htbp]
    \centering
    \begin{tabular}{@{}lcccc@{}}
    \hline
    Category & \# Parents & Recursive & Alt. rules\\ 
    \hline
    CHAIN      & 1 & No & No \\
    CHAIN REC. & 1 & Yes & No \\
    RDG        & 1 -- * & No & No \\ 
    RDG REC.   & 1 -- * & Yes & No \\
    DRDG       & 1 -- * & No & Yes \\
    DRDG REC.  & 1 -- * & Yes & Yes \\
    MIXED      & 1 -- * & Yes & Yes \\
    \hline
    \end{tabular}
    \caption{Characteristics for each dataset category. \textit{\#Parents} refers to the number of times a predicate appears as the head of a rule in the context of rule learning and logical induction. It indicates the number of rules through which each rule can deduce relevant facts. \textit{Recursive} refers to whether a predicate in the head of a rule can also occur in the body of the same rule. \textit{Alt. rules} indicate whether a predicate can be deduced by alternative rules.}
    \label{tab:complexity_cls}
\end{table}

\subsection{Dataset synthesis}

In order to generate datasets for rigorous analysis, this study employed the RuDaS tool \cite{cornelio_thost_rudas} to systematically vary parameters such as noise, open-world degree, and missing data. By adjusting these factors in conjunction with the \textit{category} parameter, it is possible to ensure comprehensive coverage of different structural configurations and expressivity levels. 

For each configuration, the following settings were applied:

\begin{itemize}
    \item The minimum and maximum number of Directed Acyclic Graphs (DAGs), which refer to the range of distinct rule structures generated within each dataset, were both set to 1 (\texttt{mindags} = 1, \texttt{maxdags} = 1).
    \item Noise levels, which include the addition of irrelevant facts or the removal of relevant ones that do not support the rules, were systematically varied at intervals of 0.1, 0.2, and 0.3.
    \item The percentage of missing data (\texttt{missing}), which determines the proportion of data entries - specifically consequences - that are intentionally left out, and open world degree (\texttt{owa}) were similarly varied across 0.1, 0.2, and 0.3.
    \item The category parameter was set to cover all classes of expressivity  described in the previous section, and listed in Table~\ref{tab:complexity_cls}. The distribution of each category in the synthesised dataset is an independent hyperparameter,  discussed in the experiments section.
\end{itemize}

Further details regarding the dataset generation can be found in the supplementary material\footref{fn:arxiv_appendix}.

\section{Proposed Approach} \label{sec:method}

The proposed approach can be divided in two parts: \textit{iterative refinement} and \textit{graded evaluation}. They form a systematic evaluation loop, covering all the expressivity classes described in the previous section, for a given set of LLMs and dataset synthesis parameters. 

\subsection{Iterative Refinement} 

Consists of an iterative refinement loop that alternates between the generation of a theory by a language model and the evaluation of said theory through a formal interpreter. It is comprised of the following components, as illustrated in Figure~\ref{fig:method}:

\noindent \textbf{(a)} A language model capable of generating a theory $H$, based on background knowledge, positive and negative examples, and hypothesis search, provided as a prompt text in a logic program language. Typically an LLM with structured (e.g., code) generation capabilities.
\noindent \textbf{(b)} A logic program interpreter. We use Prolog \cite{warren2023prolog} as the logic program language.
\noindent \textbf{(c)} A prompt generation component, that interleaves logical programming language with natural language queries designed to drive the theory induction responses. The logical programming language expresses background knowledge and the relevant outputs of the program interpreter.
\noindent \textbf{(d)} An evaluation module, that uses the logic program interpreter to execute the generated theory $H$ as logical rules, and computes a set of evaluation metrics.

\noindent Given: \textit{background knowledge (BK)}, \textit{positive examples ($E^+$)}, \textit{negative examples ($E^-$)}, and assuming a language model $\mathcal{LM}$ which can find a hypothesis $H$ (a set of logical clauses) such that it satisfies the conditions of completeness (for every example $e \in E^+$, $H \cup BK \models e$) and consistency (For every example $e \in E^-$, $H \cup BK \not\models e$). \\

\noindent 1. \textbf{Context Representation}: Represent the input to the language model as a combination of background knowledge and examples: $ \text{Context} = \text{encode}(BK, E^+, E^-)$. \\

\noindent 2. \textbf{Theory Generation}: From a background knowledge set of clauses sampled from a knowledge base dataset, including positive and negative examples, a prompt is created for the LM to induce a theory as Prolog code, i.e. using the language model to generate a set of logical clauses (hypothesis $H$): $H = \mathcal{LM}(\text{Theory Prompt + Context})$. \\

\noindent 3. \textbf{Evaluation of Hypothesis}: Checking for the completeness and consistency conditions:

\noindent \textit{True Positives (TP):} The number of positive examples correctly entailed by the hypothesis. \\
    $TP = \left| \{ e \in E^+ \mid BK \cup H \models e \} \right|$ \\
    
\noindent \textit{False Positives (FP):} The number of negative examples incorrectly entailed by the hypothesis. \\
    $FP = \left| \{ e \in E^- \mid BK \cup H \models e \} \right|$ \\

\noindent \textit{False Negatives (FN):} The number of positive examples not entailed by the hypothesis. \\
   $FN = \left| \{ e \in E^+ \mid BK \cup H \not\models e \} \right|$ \\

\noindent \textit{True Negatives (TN):} The number of negative examples correctly not entailed by the hypothesis. \\
  $TN = \left| \{ e \in E^- \mid BK \cup H \not\models e \} \right|$ \\

\noindent from which accuracy (ACC) ($\frac{TP + TN}{TP + TN + FP + FN}$), precision (P) ($\frac{TP}{TP + FP}$), recall (R) ($\text{Recall} = \frac{TP}{TP + FN}$) and F1-score (F1) ($ 2 \cdot \frac{\text{Precision} \cdot \text{Recall}}{\text{Precision} + \text{Recall}}$) can be generated. \\

\noindent 4. \textbf{Theory refinement}: Following an initial evaluation, the LM is tasked to refine the induced theory iteratively. Each refinement round involves adjusting the theory based on feedback from the Prolog interpreter validation. The refinement aims to improve the theory's performance by addressing misclassifications and enhancing its predictive capabilities. If $H$ does not satisfy completeness and consistency, update the input representation based on feedback and generate a new hypothesis using the language model, given the \textit{Feedback Context} $\leftarrow \{FP, FN, P, R, F1, ACC\}$ and the final prompt input \textit{Input} $\leftarrow$ (Refinement Prompt + Context + Feedback Context): $H \leftarrow \mathcal{LM}(\text{Input})$.

The main loop of the algorithm continues until the evaluation metrics meet the defined thresholds or the maximum number of iterations is reached. In each iteration, the language model generates a theory based on the current prompt. This generated theory is then evaluated using the logic program interpreter, in our case Prolog, resulting in a validation set. Evaluation metrics are computed from these validation results and stored. Based on the feedback from the validation, a new prompt is generated by incorporating the initial knowledge base sample, the current theory, and the validation feedback. Our approach removes recursive rules from the LLM-induced theory before evaluation. The refinement loop is summarised in Algorithm~\ref{alg:llm_method}. The process starts by sampling facts from the knowledge base dataset to create $kb$. An initial prompt is then generated using these sampled facts, denoted as $prompt$. 

\noindent 5. \textbf{Termination}: The process continues iteratively until a maximum number of iterations is reached.

\begin{algorithm}[ht!]
  \caption{Iterative LM theory refinement} \label{alg:llm_method}
  \begin{algorithmic}
      \State \textbf{Define:} $KB$ as the background knowledge dataset.
      \State \textbf{Define:} $PGen$ as the prompt generator.
      \State \textbf{Define:} $Exs$ as the positive and negative examples.
      \State \textbf{Define:} $LM$ as the language model.
      \State \textbf{Define:} $PL$ as the logic program interpreter.
      \State \textbf{Define:} $Eval$ as the evaluation module.
      \State \textbf{Define:} $M$ as the evaluation metrics set.
      \State \textbf{Define:} $Max_{iter}$ as the maximum number of iterations.
      \State \textbf{Define:} $MT_{tresh} \gets Map:M\rightarrow \mathbb{R}$ as the evaluation metrics treshold.

      \State ~

      \State \textbf{Let} $prompt \gets PGen(KB, Examples)$ 
      \State \textbf{Let} $iter \gets 0$
      \State \textbf{Let} $results \gets Map: M \rightarrow \mathbb{R}$

      \While{$(\exists m \in results: results[m] < MT_{tresh}[m]$) $\land$ ($iter < Max_{iter})$}
        \State $theory \gets LM(prompt)$
        \State $results \gets Eval(Examples)$
        \State $prompt \gets PGen(KB, theory, Exs)$
        \State $iter \gets iter + 1$
      \EndWhile
  \end{algorithmic}
\end{algorithm}

\subsection{Graded evaluation} 

A synthetic data generator is used to control for the input parameters of the ruleset expressive power, namely: categorical distribution (CHAIN, RDG, DRDG, etc.), background knowledge, positive examples and negative examples as well as the amount of noise introduced within the dataset. \\

\noindent 1. \textbf{Categorised Learning Sets}: Consisting of $C$: set of ruleset expressivity classes (e.g., CHAIN, RDG, DRDG, etc.), $N$: set of noise levels, $S$: number of samples per combination of $C$ and $N$. For each $c \in C$ and $n \in N$, generate $S$ datasets $\{D_{c,n,i} \mid i = 1, \ldots, S\}$ where each dataset $D_{c,n,i}$ includes:
        \[
        D_{c,n,i} = (\text{BK}_{c,n,i}, E^+_{c,n,i}, E^-_{c,n,i}, \text{noise}_{c,n})
        \]
        
\noindent 2. \textbf{Hypothesis Generation and Evaluation}: For each dataset $D_{c,n,i}$, use a learning algorithm to generate a hypothesis $H_{c,n,i}$, tracking the F1 score $F_{c,n,i}$ and processing time $T_{c,n,i}$ at each iteration and recording the best F1 score $\text{F1}_{c,n,i}$ and corresponding processing time $\text{Time}_{c,n,i}$:
        \[
        \text{F1}_{c,n,i} = \max(F_{c,n,i})
        \]
        \[
        \text{Time}_{c,n,i} = \text{time until } \max(F_{c,n,i})
        \]

\noindent 3. \textbf{Aggregation}: The information is then aggregated by complexity category and noise level for all the samples, averaging times and F1 scores, to obtain the complete graded evaluation statistics. For each combination of $c \in C$ and $n \in N$, compute the average F1 score and average processing time:
        \[
        \overline{\text{F1}}_{c,n} = \frac{1}{S} \sum_{i=1}^{S} \text{F1}_{c,n,i}
        \]
        \[
        \overline{\text{Time}}_{c,n} = \frac{1}{S} \sum_{i=1}^{S} \text{Time}_{c,n,i}
        \]

\section{Experiments} \label{sec:exp}

In order to answer the research questions, a set of experiments was elaborated to systematically analyse the theory inducing capabilities of a set of most popular open-source LLMs and two versions of GPT, with the proposed approach, having the state-of-the-art ILP system \textit{Popper}~\cite{cropper2021learning} as a baseline. The tests covered all data categories discussed on the section ``\textit{Inductive Learning, Expressive Power \& Datasets}'', allowing a graded analysis w.r.t. the expected level of induction expressivity and tolerated noise.

\subsection{Experimental Setup \& Dataset}

For each data category, five datasets were generated using \textit{RuDaS}~\cite{cornelio_thost_rudas}. The size of each dataset was set to $XS$ ($min = 50$, $max = 100$, $support = 3$), and noise, missing, and open world were all set to 0.1, then all set to 0.2, and finally all set to 0.3. This resulted in 105 datasets in total, with 35 datasets for each rate. Subsequently, two methods are used to induce a theory for each dataset: (1) employing \textit{Popper}, with NuWLS \cite{Chu_Cai_Luo_2023} and WMaxCDCL, varying its time limit parameter from 10 to 800 seconds; (2) applying the proposed iterative LM theory refinement method (Section ``\textit{Proposed Approach}''), with parameters $Max_{iter} = 4$ and $MT_{thresh} = 1.0$. Three different LLM models were used for (2): Open AI\footnote{\url{https://openai.com/}}'s model \textit{GPT-4o}\footnote{\url{https://openai.com/index/hello-gpt-4o}}, Mistral AI\footnote{\url{https://mistral.ai/}}'s \textit{Mixtral-8x7B}\footnote{\url{https://huggingface.co/mistralai/Mixtral-8x7B-Instruct-v0.1}}~\cite{jiang2023mistral}, and Google's \textit{Gemma}\footnote{\url{https://huggingface.co/google/gemma-7b-it}}~\cite{team2023gemini}.

Table \ref{tab:datastats} presents a comprehensive overview of statistical metrics pertaining to each category of data, in order of complexity (except \textit{MIXED}).

\begin{table}[htbp]
    \centering
    \label{tab:my-table}
    \begin{tabular}{@{}llll@{}}
    \hline
    Categories                        & Facts       & Positive & Negative \\ 
    \hline
    CHAIN & 67.6 $\pm$ 5.4        & 23.6 $\pm$ 6.5              & 4.8 $\pm$ 2.1               \\
    CHAIN REC. & 54.2 $\pm$ 14.1       & 19.6 $\pm$ 7.8              & 4.4 $\pm$ 2.5               \\
    RDG & 60.2 $\pm$ 9.8        & 18.6 $\pm$ 12.3             & 4.2 $\pm$ 3.4               \\ 
    RDG REC. & 63.2 $\pm$ 9.0        & 16.6 $\pm$ 4.7              & 3.4 $\pm$ 1.3               \\
    DRDG    & 61.2 $\pm$ 14.1       & 31.0 $\pm$ 26.0             & 8.0 $\pm$ 8.2               \\
    DRDG REC. & 54.2 $\pm$ 12.5       & 34.0 $\pm$ 22.2             & 9.0 $\pm$ 6.2               \\
    MIXED & 54.4 $\pm$ 18.1       & 24.2 $\pm$ 12.3             & 4.8 $\pm$ 2.7               \\
    \hline
    \end{tabular}
    \caption{Statistics for each dataset category. A detailed description of each can be found in Section ``\textit{Inductive Learning, Expressive Power \& Datasets}''. The mean $\pm$ standard deviation (mean $\pm$ std) for each category is provided}
    \label{tab:datastats}
\end{table}

We computed the average F1-score for each category, taking into account the level of noise, open world scenarios, and missing facts. The mean values reported are based on the results obtained from the theory that was generated from the train set and evaluated on the test set.












The experiment used the OpenAI service for GPT models. For Popper, Llama3-8B-Instruct, Gemma-7B-It and Mixtral-8x7B-Instruct-v0.1, it was conducted on a computer with an Intel(R) Xeon(R) Gold 5217 CPU @ 3.00GHz, 188GB RAM, and 2x NVIDIA RTX A6000 (48GB VRAM) GPUs. The software used was CUDA 12.3, PyTorch 2.2.2, and Transformers 4.41.2. Prompt templates used were included in the supplementary material\footref{fn:arxiv_appendix}. 


\subsection{Results \& Discussion}

Overall results for F1 are presented in Figure~\ref{fig:f1}. We additionally report on processing times as a measure of practical interest in Figure~\ref{fig:time}. We present the values obtained in detail in the supplementary material\footref{fn:arxiv_appendix} (Tables 4 and 5) 
\textit{Gemma-7B-It} results are not included as it failed to generate valid theories. The results reveal significant insights into LLM capabilities and limitations w.r.t. theory induction, which are summarised as follows:

\begin{figure}[htb!]
        \centering
        \includegraphics[width=1.00\linewidth]{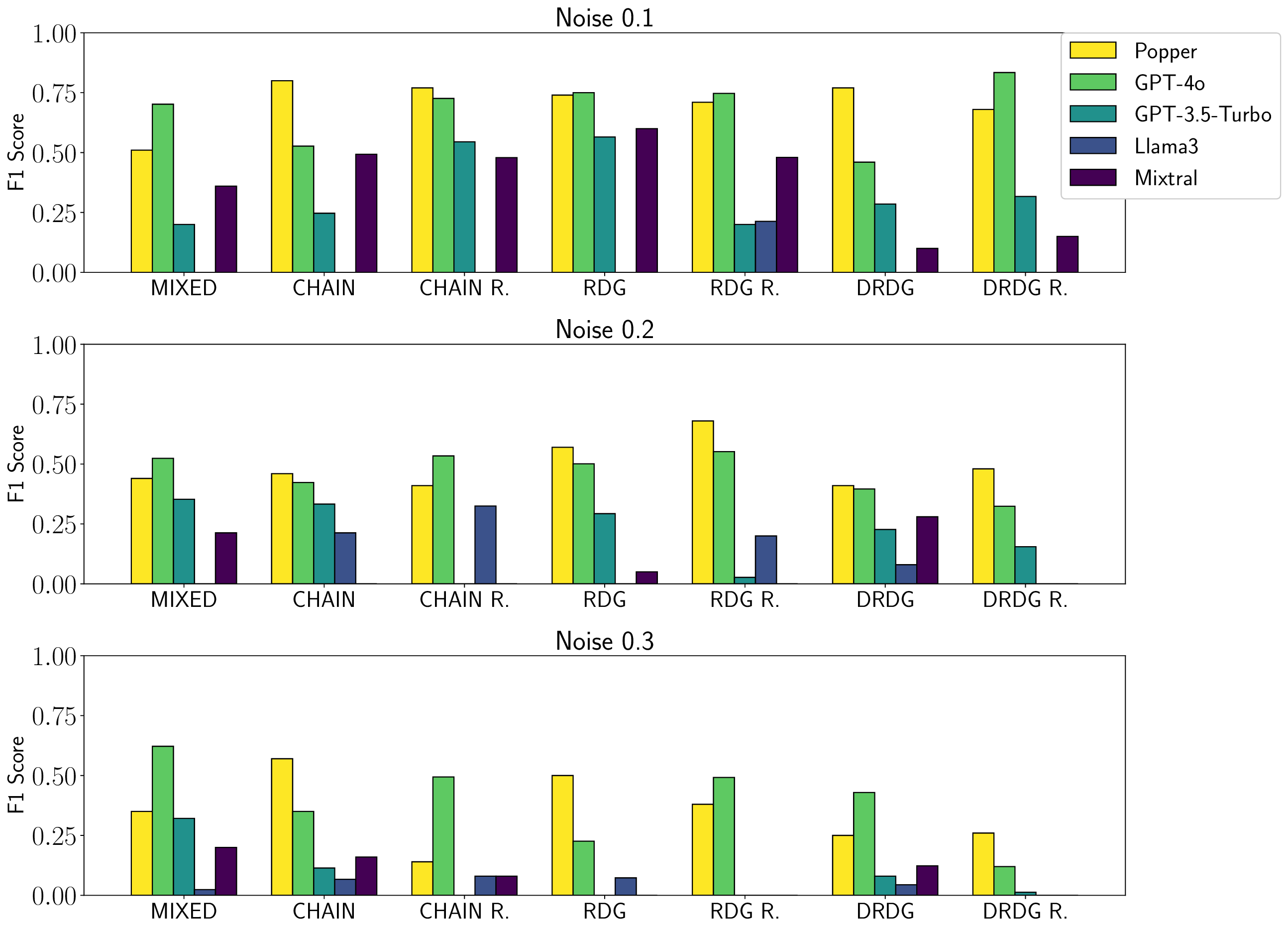}
        \caption{F1 score trends across categories. Different models (GPT-4o, Llama3 8b instruct, Popper, and Mixtral-8x7B-Instruct-v0.1) under varying noise levels and categories reveal distinct performance patterns. GPT-4o demonstrates stable accuracy yet sensitivity to noise, particularly in complex rule-based categories like RDG and DRDG. Mixtral-8x7B-Instruct-v0.1 exhibits mixed results with notable variability across categories particularly in more complex tasks. Llama3 8b instruct delivers lower scores, indicating challenges in reasoning and theory generation.}
        \label{fig:f1}
\end{figure}

\begin{figure}[htb!]
        \centering
        \includegraphics[width=1.0\linewidth]{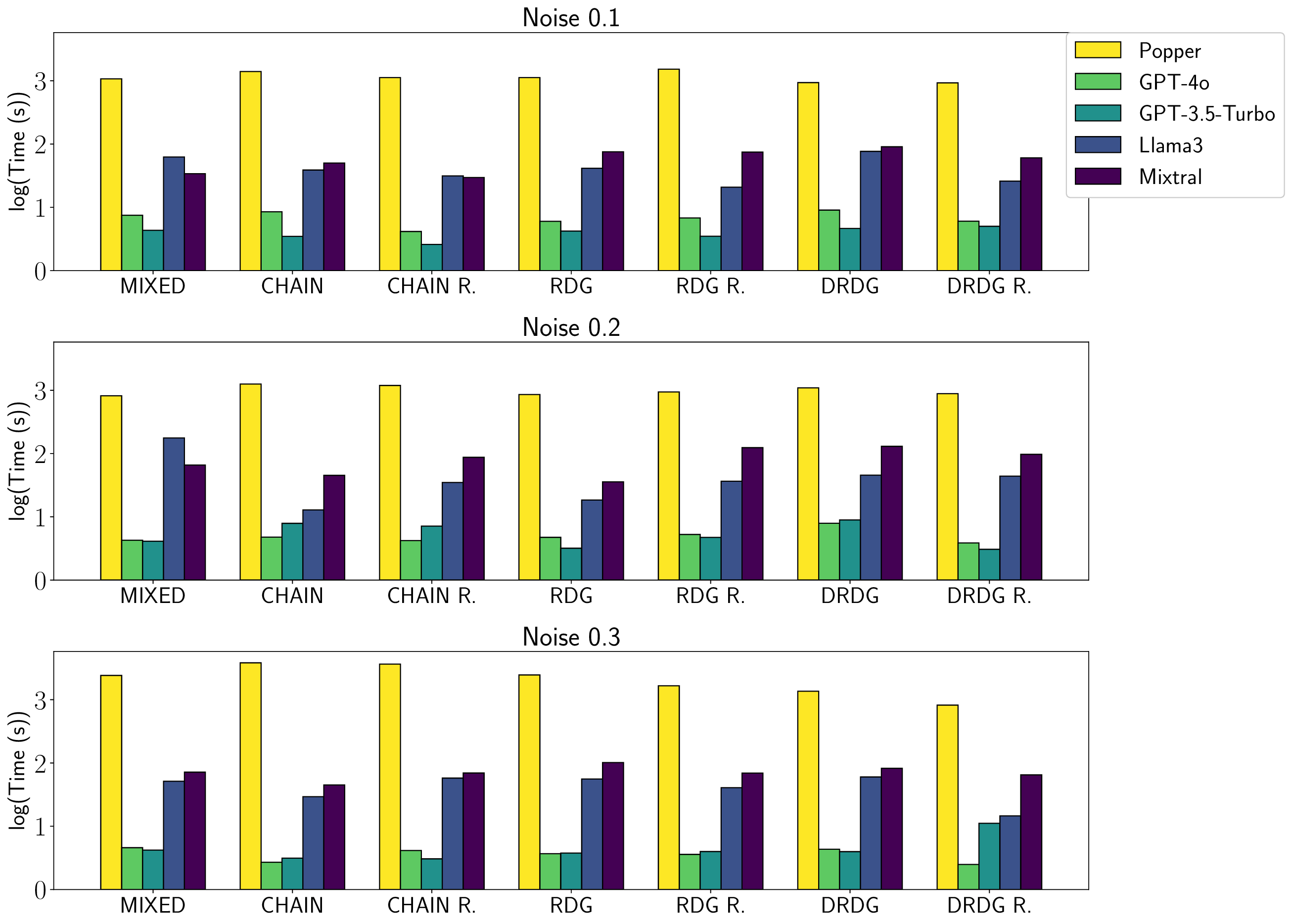}
        \caption{Performance on time consumption trends across categories using a logarithmic scale. The data consistently shows that LLM outperforms Popper in all intervals. The results however do not represent a measure of efficiency, as the computational resources employed are vastly different across methods.}
        \label{fig:time}
\end{figure}

\paragraph{LLMs can achieve competitive performance against the baseline, specially at higher noise levels.}
The larger scale models (GPT3.5, 4) demonstrate more resilience to noise and consistent F1 across the different categories, as indicated in Figure~\ref{fig:f1}, with an average F1-score difference of $\pm0.25$ against Popper. This answers the overall quality part of \textbf{RQ1}.

\paragraph{Inducing theories on long relation chains is the major obstacle for LLMs, rather than the expressivity level.}
With the CHAIN category being the least expressive and one of the most consistently solved by Popper, but none of the tested LLMs was able to overcome the baseline performance at all noise levels (Figure~\ref{fig:f1} \textit{CHAIN} category). This suggests that such models have a limited capacity of tracking long relationship chains of independent predicates. This addresses a part of the answer for \textbf{RQ2}.

\paragraph{Increasing iteration limits does not monotonically improve results for LLMs.}
Upon increasing the iteration limits from 1 to 4, it was found that the metrics can either increase or decrease non-monotonically. No significant overall performance improvements were observed beyond 3 iterations. Thus $Max_{iter}$ was set to 4 and the iteration with the best accuracy is taken as the final result.

\paragraph{Performance is remarkably lower on more expressive rule sets at moderate noise levels.}
Responses for more expressive categories, such as RDG and DRDG display higher variance and LLM hallucination artifacts, such as valid rules containing predicates that do not exist in the rule set. We present an error analysis in the section ``\textit{Error analysis}''. For instance, a comparison of the results for the RDG category generated by GPT-4o under noise levels set to 0.1 and 0.3 reveals a significant decline in performance, with the F1-score dropping from 0.75 to 0.22. A comparable pattern is observed with GPT-3.5 Turbo for RDG and DRDG and Mixtral RDG in the presence of elevated noise levels, with GPT-3.5 Turbo scores going from 0.56 to 0.0, 0.28 to 0.08, and Mixtral-8x7B going from 0.60 to 0.0. This complements the answer to \textbf{RQ2}. Further details are included in the supplementary material\footref{fn:arxiv_appendix}.

\paragraph{Induction capability varies substantially across models.}
Using the same inputs resulted in vastly different responses from different models, suggesting critical influence from model size in parameters. Figure \ref{fig:f1} illustrates this: when comparing GPT-4o, Mixtral-8x7B and Llama3 at noise levels set to 0.1 and 0.3 respectively, the consistency in generating a valid theory correlates to their relative size.

At a noise level of 0.1, GPT-4o's F1 score is almost twice that of the GPT-3.5-Turbo in average, and at a noise level of 0.3, the difference increases to a ratio of 4, indicating substantially higher noise resiliency. The performance gap is more pronounced when comparing with Llama3-8B, where GPT-4o F1 score is 21 times higher at the lowest noise setting.


Mixtral-8x7B-It-v0.1 performs similarly to GPT-3.5-Turbo at lower noise levels, scoring 13.4\% higher in average at 0.1 noise. However, its performance becomes less stable at higher noise levels. It consistently outperforms Llama3-8B-it, at 0.1 noise, with a F1-score 11 times higher in average.

\paragraph{Model size does not correlate to noise resilience}
Despite being able to achieve higher scores than GPT-3.5 and Mixtral-8x7B in some of the tests (e.g., RDG-R $@noise=0.1$, CHAIN-R $@noise=0.2$) and scoring higher on intermediate noise than on low noise, Llama3-8B did not consistently generate valid theories. On the other hand, Mixtral-8x7B, a much larger model, is particularly susceptible to noise, with an average F1-score drop of over 0.8 from $noise = 0.1$ to $noise = 0.2$ and a monotonic performance reduction with the increase of the noise level.\\

\noindent Regarding the parameterisation of each method, some higher-level observations on the trade-off between time and inference quality can be summarised as follows:

\paragraph{Computational scalability allow LLMs to operate at substantially lower times.}
While Popper consists of a combination of serial algorithms, the parallel nature of transformer-based LLMs allows them to operate at times about 3 orders of magnitude lower, given enough computational resources. This can be observed in Figure \ref{fig:time}, for all  classes and all tested noise intervals.

\subsection{Error analysis} \label{sec:err_analysis}
The errors found in the evaluation of the generated theories could be separated in two categories: \textit{syntactic} and \textit{logical}.

\noindent \textit{Syntactic} errors occur when the generated response does not match the logic programming language grammar. For example, the following response: 

\begin{minted}{prolog}
theory :-
    p(X, Y), pos(p0(X, Y)) - positive.
    p(X, Y), neg(p0(X, Y)) - negative.
    \+ p(X, Y), pos(p0(X, Y)) - false.
    \+ p(X, Y), neg(p0(X, Y)) - true.
\end{minted}

\noindent is not valid Prolog and will fail evaluation.\\

\noindent \textit{Logical} errors occur when the generated response has correct grammar, but cannot be induced from the examples. Consider the following Prolog theory:

\begin{minted}{prolog}
theory :-
    p(X, Y) :- p1(X, Y); p3(X, Y); 
    p4(X, Y); p7(X, Y); p8(X, Y); 
    p0(X, Y),
    not neg(p(X, Y)),
    (pos(p(X, Y)) - true; fail).
\end{minted}

The response contains the head of the clause "theory," as well as the predicates "p" and "pos", which do not exist in the BK. Table~\ref{tab:err_quant} presents a distribution of error categories for the analysed models. A more detailed analysis of the models outputs is included in the supplementary material\footref{fn:arxiv_appendix}.

\begin{table}[htbp]
    \centering
    \begin{tabular}{@{}lcc@{}}
    \hline
    Model         & \# Syntactic & Logical\\ 
    \hline
    GPT-4o         & 0\% & 100\%\\
    GPT3.5        & 0\% & 100\%\\
    Llama3-8B     & 46\% & 54\%\\ 
    Mixtral-8x7B  & 20\% & 80\%\\
    Gemma-7B-it  & 100\% & 0\%\\
    \hline
    \end{tabular}
    \caption{Error distribution for each of the evaluated models. Gemma-7B-it did not produce valid Prolog.}
    \label{tab:err_quant}
\end{table}

\section{Related Work} \label{sec:related}

Neural symbolic computation combines neural networks with symbolic methods to enhance AI reasoning capabilities. Yang and Cohen~\shortcite{yang2017differentiable} introduced Neural Logic Programming, An end-to-end differentiable model integrating neural networks with logic programming. Within the LLM-Symbolic space, Wan et al.~\shortcite{DBLP:journals/corr/abs-2401-00757} developed LogicAsker, which evaluates and improves LLMs' logical reasoning using propositional and predicate logic. It identifies reasoning failures and enhances capabilities through in-context learning. Within the context of symbolic toolformers over LLMs, Quan et al.~\shortcite{quan2024enhancing, quan2024verification} proposed methods of improving explanatory reasoning with the support of formal iterative cycles using both logical solvers and theorem provers for supporting more controlled step-wise reasoning.

Despite these advancements at the interface of LLM-based reasoning and formal controls, it is unclear the extent and the conditions in which LLMs can perform formal reasoning \cite{huang-chang-2023-towards}. Sinha et al.~\shortcite{sinha-etal-2019-clutrr} introduced CLUTRR, a benchmark assessing LLMs' structural learning by inferring kinship relations in stories, requiring relationship extraction and logical rule inference. Zhu et al.~\shortcite{zhu2024large} proposed the Hypotheses-to-Theories (HtT) framework to improve LLM reasoning by learning explicit rules in two stages: generating and verifying rules (induction) and using the obtained rule library for reasoning (deduction). HtT enhances relational and numerical reasoning and concept learning.

Madaan et al.~\shortcite{madaan2023selfrefine} introduces a novel technique for improving machine learning models through iterative refinement. This approach allows models to improve their performance by continuously evaluating and adjusting their predictions based on self-generated feedback. By critiquing their own outputs, models can identify errors and make corrections over successive iterations, leading to increased accuracy and robustness across different tasks. Our work builds upon this approach by employing an ILP solver to evaluate and then using its output as critique for the refinement process.

In a related study \cite{dziri2023faith}, the authors investigate the limitations of transformer models in handling composition tasks. Their results show that, despite their strengths, transformers face significant challenges in dealing with compositionality, which involves understanding and generating complex structures from simpler components. This limitation highlights the need for innovative approaches, such as self-refining, to further enhance the capabilities of machine learning models.

In contrast, our work focuses on the still under-explored area of assessing and controlling inductive learning/inference capabilities of LLMs. These contributions integrate LLMs and formal logic for robust theory induction and allows a graded analysis of LLM capabilities, with respect to theory induction complexity.  

\section{Conclusion} \label{sec:conclusion}

In this study we thoroughly investigate the integration of state-of-the-art formal theory induction within the context of large language models (LLMs), aiming to elucidate the extent in which LLMs can systematically perform inductive learning for theories spanning across different expressivity levels. At the heart of this exploration lies the recognition of relational data's inherent semantic depth, stemming from its symbolic representations. The empirical results presented here have indicated the ability of LLMs to address inductive learning tasks, with the largest LLMs achieving competitive results against the algorithmic SOTA with better tolerance to higher noise levels, which can be attributed to their semantic flexibility. This flexibility however has certain limitations, as we found that tested language models are more limited by their capacity of tracking long relationship chains of independent predicates than by the dependency complexity of the rule sets (disjunctiveness, recursivity).


As future work we plan to use larger datasets to test the scalability of the proposed approach, allowing the assessment of the performance of LLMs across a broader range of domain-specific and agnostic scenarios. Additionally, we will investigate the impact of LLMs for bootstrapping the initial generation of theories.

\section{Limitations}

While the proposed evaluation methodology aims to cover a wide range of logic theory induction expressivity levels, it is limited in its resolution to the categories specified by \cite{cornelio_thost_rudas}, and does not quantify other ruleset characteristics, such as \textit{workspace size} or \textit{unification rate} in the case of Prolog \cite{dikovsky1993computational}.

The methodology compares all models under the same inputs. Therefore it is not concerned with extracting maximum performance of any given model, but obtaining a relative assessment of their fundamental capabilities. This means that the empirical analysis reported scores should not be taken as a measure of SOTA performance.

Furthermore, the time gains demonstrated in the experiments are presented as an achievable result, conditioned to the combination of software and hardware indicated in the paper and the services provided by third-parties (e.g., OpenAI). They should not be interpreted as a measure of computational efficiency.


\section{Acknowledgements}
This work was partially funded by the Swiss National Science Foundation (SNSF) project NeuMath (200021\_204617)\footnote{\url{https://data.snf.ch/grants/grant/204617}}, by the CRUK National Biomarker Centre, and supported by the Manchester Experimental Cancer Medicine Centre and the NIHR Manchester Biomedical Research Centre.


\bibliography{references}

\begin{thebibliography}{20}
\providecommand{\natexlab}[1]{#1}

\bibitem[{Brown et~al.(2020)Brown, Mann, Ryder, Subbiah, Kaplan, Dhariwal, Neelakantan, Shyam, Sastry, Askell, Agarwal, Herbert-Voss, Krueger, Henighan, Child, Ramesh, Ziegler, Wu, Winter, Hesse, Chen, Sigler, Litwin, Gray, Chess, Clark, Berner, McCandlish, Radford, Sutskever, and Amodei}]{10.5555/3495724.3495883}
Brown, T.~B.; Mann, B.; Ryder, N.; Subbiah, M.; Kaplan, J.; Dhariwal, P.; Neelakantan, A.; Shyam, P.; Sastry, G.; Askell, A.; Agarwal, S.; Herbert-Voss, A.; Krueger, G.; Henighan, T.; Child, R.; Ramesh, A.; Ziegler, D.~M.; Wu, J.; Winter, C.; Hesse, C.; Chen, M.; Sigler, E.; Litwin, M.; Gray, S.; Chess, B.; Clark, J.; Berner, C.; McCandlish, S.; Radford, A.; Sutskever, I.; and Amodei, D. 2020.
\newblock Language models are few-shot learners.
\newblock In \emph{Proceedings of the 34th International Conference on Neural Information Processing Systems}, NIPS '20. Red Hook, NY, USA: Curran Associates Inc.
\newblock ISBN 9781713829546.

\bibitem[{Chu, Cai, and Luo(2023)}]{Chu_Cai_Luo_2023}
Chu, Y.; Cai, S.; and Luo, C. 2023.
\newblock NuWLS: Improving Local Search for (Weighted) Partial MaxSAT by New Weighting Techniques.
\newblock \emph{Proceedings of the AAAI Conference on Artificial Intelligence}, 37(4): 3915--3923.

\bibitem[{Cornelio and Thost(2021)}]{cornelio_thost_rudas}
Cornelio, C.; and Thost, V. 2021.
\newblock Synthetic Datasets and Evaluation Tools for Inductive Neural Reasoning.
\newblock In \emph{Proceedings of the {30th} International Conference on Inductive Logic Programming, ILP2020-21 @ IJCLR}.

\bibitem[{Cropper and Morel(2021)}]{cropper2021learning}
Cropper, A.; and Morel, R. 2021.
\newblock Learning programs by learning from failures.
\newblock \emph{Machine Learning}, 110(4): 801--856.

\bibitem[{Dikovsky(1993)}]{dikovsky1993computational}
Dikovsky, A.~J. 1993.
\newblock On computational complexity of Prolog programs.
\newblock \emph{Theoretical Computer Science}, 119(1): 63--102.

\bibitem[{Dziri et~al.(2023)Dziri, Lu, Sclar, Li, Jian, Lin, West, Bhagavatula, Bras, Hwang et~al.}]{dziri2023faith}
Dziri, N.; Lu, X.; Sclar, M.; Li, X.~L.; Jian, L.; Lin, B.~Y.; West, P.; Bhagavatula, C.; Bras, R.~L.; Hwang, J.~D.; et~al. 2023.
\newblock Faith and Fate: Limits of Transformers on Compositionality.
\newblock \emph{arXiv preprint arXiv:2305.18654}.

\bibitem[{Huang and Chang(2023)}]{huang-chang-2023-towards}
Huang, J.; and Chang, K. C.-C. 2023.
\newblock Towards Reasoning in Large Language Models: A Survey.
\newblock In Rogers, A.; Boyd-Graber, J.; and Okazaki, N., eds., \emph{Findings of the Association for Computational Linguistics: ACL 2023}, 1049--1065. Toronto, Canada: Association for Computational Linguistics.

\bibitem[{Jiang et~al.(2023)Jiang, Sablayrolles, Mensch, Bamford, Chaplot, Casas, Bressand, Lengyel, Lample, Saulnier et~al.}]{jiang2023mistral}
Jiang, A.~Q.; Sablayrolles, A.; Mensch, A.; Bamford, C.; Chaplot, D.~S.; Casas, D. d.~l.; Bressand, F.; Lengyel, G.; Lample, G.; Saulnier, L.; et~al. 2023.
\newblock Mistral 7B.
\newblock \emph{arXiv preprint arXiv:2310.06825}.

\bibitem[{Madaan et~al.(2023)Madaan, Tandon, Gupta, Hallinan, Gao, Wiegreffe, Alon, Dziri, Prabhumoye, Yang, Welleck, Majumder, Gupta, Yazdanbakhsh, and Clark}]{madaan2023selfrefine}
Madaan, A.; Tandon, N.; Gupta, P.; Hallinan, S.; Gao, L.; Wiegreffe, S.; Alon, U.; Dziri, N.; Prabhumoye, S.; Yang, Y.; Welleck, S.; Majumder, B.~P.; Gupta, S.; Yazdanbakhsh, A.; and Clark, P. 2023.
\newblock Self-Refine: Iterative Refinement with Self-Feedback.
\newblock arXiv:2303.17651.

\bibitem[{Muggleton(1991)}]{ilpmugg}
Muggleton, S. 1991.
\newblock Inductive logic programming.
\newblock \emph{New generation computing}, 8: 295--318.

\bibitem[{Nienhuys-Cheng and de~Wolf(1997)}]{ilpwolf}
Nienhuys-Cheng, S.-H.; and de~Wolf, R. 1997.
\newblock \emph{What is inductive logic programming?}
\newblock Springer.

\bibitem[{Quan et~al.(2024{\natexlab{a}})Quan, Valentino, Dennis, and Freitas}]{quan2024enhancing}
Quan, X.; Valentino, M.; Dennis, L.~A.; and Freitas, A. 2024{\natexlab{a}}.
\newblock Enhancing Ethical Explanations of Large Language Models through Iterative Symbolic Refinement.
\newblock arXiv:2402.00745.

\bibitem[{Quan et~al.(2024{\natexlab{b}})Quan, Valentino, Dennis, and Freitas}]{quan2024verification}
Quan, X.; Valentino, M.; Dennis, L.~A.; and Freitas, A. 2024{\natexlab{b}}.
\newblock Verification and Refinement of Natural Language Explanations through LLM-Symbolic Theorem Proving.
\newblock arXiv:2405.01379.

\bibitem[{Radford et~al.(2019)Radford, Wu, Child, Luan, Amodei, Sutskever et~al.}]{radford2019language}
Radford, A.; Wu, J.; Child, R.; Luan, D.; Amodei, D.; Sutskever, I.; et~al. 2019.
\newblock Language models are unsupervised multitask learners.
\newblock \emph{OpenAI blog}, 1(8): 9.

\bibitem[{Sinha et~al.(2019)Sinha, Sodhani, Dong, Pineau, and Hamilton}]{sinha-etal-2019-clutrr}
Sinha, K.; Sodhani, S.; Dong, J.; Pineau, J.; and Hamilton, W.~L. 2019.
\newblock {CLUTRR}: A Diagnostic Benchmark for Inductive Reasoning from Text.
\newblock In Inui, K.; Jiang, J.; Ng, V.; and Wan, X., eds., \emph{Proceedings of the 2019 Conference on Empirical Methods in Natural Language Processing and the 9th International Joint Conference on Natural Language Processing (EMNLP-IJCNLP)}, 4506--4515. Hong Kong, China: Association for Computational Linguistics.

\bibitem[{Team et~al.(2023)Team, Anil, Borgeaud, Wu, Alayrac, Yu, Soricut, Schalkwyk, Dai, Hauth et~al.}]{team2023gemini}
Team, G.; Anil, R.; Borgeaud, S.; Wu, Y.; Alayrac, J.-B.; Yu, J.; Soricut, R.; Schalkwyk, J.; Dai, A.~M.; Hauth, A.; et~al. 2023.
\newblock Gemini: a family of highly capable multimodal models.
\newblock \emph{arXiv preprint arXiv:2312.11805}.

\bibitem[{Wan et~al.(2024)Wan, Wang, Yang, Yuan, Huang, He, Jiao, and Lyu}]{DBLP:journals/corr/abs-2401-00757}
Wan, Y.; Wang, W.; Yang, Y.; Yuan, Y.; Huang, J.; He, P.; Jiao, W.; and Lyu, M.~R. 2024.
\newblock A {\&} {B} == {B} {\&} {A:} Triggering Logical Reasoning Failures in Large Language Models.
\newblock \emph{CoRR}, abs/2401.00757.

\bibitem[{Warren et~al.(2023)Warren, Dahl, Eiter, Hermenegildo, Kowalski, and Rossi}]{warren2023prolog}
Warren, D.~S.; Dahl, V.; Eiter, T.; Hermenegildo, M.~V.; Kowalski, R.; and Rossi, F. 2023.
\newblock \emph{Prolog: The Next 50 Years}, volume 13900.
\newblock Springer Nature.

\bibitem[{Yang, Yang, and Cohen(2017)}]{yang2017differentiable}
Yang, F.; Yang, Z.; and Cohen, W.~W. 2017.
\newblock Differentiable learning of logical rules for knowledge base reasoning.
\newblock \emph{Advances in neural information processing systems}, 30.

\bibitem[{Zhu et~al.(2024)Zhu, Xue, Chen, Zhou, Tang, Schuurmans, and Dai}]{zhu2024large}
Zhu, Z.; Xue, Y.; Chen, X.; Zhou, D.; Tang, J.; Schuurmans, D.; and Dai, H. 2024.
\newblock Large Language Models can Learn Rules.
\newblock arXiv:2310.07064.

\end{thebibliography}

\clearpage

\appendix

\section*{Appendices}

\section{Further theoretical background}


\subsection{Detailed Complexity Classes} \label{sec:complexity_cls}
\noindent \textbf{Category Chain.} In this category, every rule, except the root, deduces facts relevant to precisely one other rule. Essentially, each node has at most one parent, and each rule is associated with at most one other rule that might infer relevant facts. Recursive rules, where the predicate in the head also occurs in the body, are exceptions, as they are relevant both for themselves and one additional rule. For example:

\begin{verbatim}
p5(X, Y) :- p0(X, Z), P2(Y, W).
p0(X, Y) :- p3(X, Z), p4(W, Y).
p3(X, Y) :- p6(X, Z), p7(W, Y).
\end{verbatim}

For example, according to Rule 1, \( p0(X, Z) \) is necessary for \( p5(X, Y) \). Therefore, satisfying \( p5(X, Y) \) requires \( p0(X, Z) \), which in turn requires \( p3(X, Z) \) and \( p4(W, Y) \). This creates a dependency chain where \( p5(X, Y) \) relies on \( p3(X, Z) \) and \( p4(W, Y) \).\\

\noindent \textbf{Category Rooted DG (RDG).} This category generalises the Chain category. Here, every rule can be relevant for several others, and each node can have multiple parent nodes. Furthermore, for each rule, there may be several other rules that might infer facts relevant for it. However, for each predicate occurring in the body of a rule, there must be at most one other rule with this predicate in the head. In other words, there are no alternative rules to derive facts relevant for a rule with respect to a specific body atom. For example:

\begin{verbatim}
p0(X0,X1) :- p1(X1,X2),p3(X0,X1).
p3(X0,X1) :- p8(X0,X1),p6(X0,X1).
p1(X1,X2) :- p7(X2,X1).
\end{verbatim}

In the given example, each rule has at least one child node. For instance, \( p0(X0, X1) \) has two child nodes: \( p1(X1, X2) \) and \( p3(X0, X1) \). Each predicate in the body of a rule corresponds to at most one rule with that predicate in the head. There are no alternative rules for deriving facts related to a specific body atom. For example, \( p1(X1, X2) \) appears in the body of \( p0(X0, X1) \) and only one rule has \( p1(X1, X2) \) in the head: \( p1(X1, X2) \) :- \( p7(X2, X1) \). The same applies to \( p3 \).\\



\noindent \textbf{Category Disjunctive Rooted DG (DRDG).} Category DRDG generalises Category RDG by allowing for alternative rules represented as children of an "OR" node. For instance:
\begin{verbatim}    
p7(X0,X1) :- p5(X0,X1).
p5(X0,X1) :- p0(X0,X1).
p5(X0,X1) :- p8(X1,X0).
\end{verbatim}

In the example, the first rule states \( p7(X0, X1) \) is true if \( p5(X0, X1) \) is true, indicating \( p7 \) depends on \( p5 \). The second rule states \( p5(X0, X1) \) is true if \( p0(X0, X1) \) is true, showing \( p5 \) depends on \( p0 \). The third rule states \( p5(X0, X1) \) is true if \( p8(X1, X0) \) is true, adding an alternative condition with swapped arguments. Thus, \( p5 \) acts as an "OR" condition in the first rule's body and the second and third rules' heads.\\

\noindent \textbf{Category Mixed.} A rule graph in this category contains connected components of different categories mentioned above. Additionally, recursion is allowed, meaning that the head of the rule may appear in the body as well.

\section{Further empirical data \& findings}
\label{sec:appendix}

\begin{table*}[ht!]
    \centering
    \small
    \begin{tabular}{@{}llllllllll@{}}
    \hline
    Category & \multicolumn{2}{c}{Noise 0.1} & \multicolumn{2}{c}{Noise 0.2} & \multicolumn{2}{c}{Noise 0.3} \\ 
    \hline
           & F1 & Time (s) & F1 & Time (s) & F1 & Time (s) \\
    \hline
    MIXED & 0.51 & 1071.53 
    & 0.44 & 817.15
    & 0.35 & 2418.88 \\
    CHAIN & 0.80 & 1397.35
    & 0.46 & 1254.33
    & 0.57 & 3824.31 \\
    CHAIN R. & 0.77 & 1123.25
    & 0.41 & 1190.14
    & 0.14 & 3646.43 \\
    RDG & 0.74 & 1122.13
    & 0.57 & 854.85
    & 0.50 & 2460.90 \\
    RDG R. & 0.71 & 1523.37
    & 0.68 & 940.50
    & 0.38 & 1659.27 \\
    DRDG & 0.77 & 934.98
    & 0.41 & 1089.47
    & 0.25 & 1363.27 \\
    DRDG R. & 0.68 & 927.30
    & 0.48 & 882.28
    & 0.26 & 820.51 \\
    \hline
    \end{tabular}
    \caption{Results for different categories with theory induced by Popper and different noise levels.}
    \label{tab:popper_res}
    \end{table*}

\begin{table*}[ht!]
    \centering
    \small
    \begin{tabular}{lcccccccc}
    \hline
    \textbf{Category} & \multicolumn{2}{c}{\textbf{GPT-4o - noise 0.1}} & \multicolumn{2}{c}{\textbf{GPT-4o - noise 0.2}} & \multicolumn{2}{c}{\textbf{GPT-4o - noise 0.3}} \\
    \hline
      & F1 (avg) & Time (s) & F1 (avg) & Time (s) & F1 (avg) & Time (s) \\
    \hline
    MIXED & 0.70 & 8.57 & 
    0.52 & 10.73 & 
    0.62 & 9.81 \\
    CHAIN & 0.52 & 8.54 & 
    0.42 & 11.05 & 
    0.35 & 8.29 \\
    CHAIN R. & 0.72 & 11.48 & 
    0.53 & 8.86 & 
    0.49 & 8.13  \\
    RDG & 0.75 & 8.80 & 
    0.50 & 11.94 & 
    0.22 & 20.16 \\
    RDG R. & 0.74 & 10.83 & 
    0.55 & 7.35 & 
    0.49 & 10.79 \\
    DRDG & 0.46 & 12.59 & 
    0.39 & 16.44 & 
    0.42 & 11.14 \\
    DRDG R. & 0.83 & 13.11 & 
    0.32 & 13.29 & 
    0.12 & 8.45 \\
    \hline
    \textbf{Category} & \multicolumn{2}{c}{\textbf{GPT-3.5-Turbo - noise 0.1}} & \multicolumn{2}{c}{\textbf{GPT-3.5-Turbo - noise 0.2}} & \multicolumn{2}{c}{\textbf{GPT-3.5-Turbo - noise 0.3}} \\
    \hline
     & F1 (avg) & Time (s) & F1 (avg) & Time (s) & F1 (avg) & Time (s) \\
    \hline
    MIXED & 0.20 & 4.32 &
    0.35 & 4.11 &
    0.32 & 4.20 \\
    CHAIN & 0.24 & 3.47 &
    0.33 & 7.88 &
    0.11 & 3.13 \\
    CHAIN R. & 0.545 & 2.592 &
    0.00 & 7.11 &
    0.00 & 3.05 \\
    RDG & 0.56 & 4.22 &
    0.29 & 3.19 &
    0.00 & 3.76 \\
    RDG R. & 0.20 & 3.48 &
    0.02 & 4.73 &
    0.00 & 3.98 \\
    DRDG & 0.28 & 4.63 &
    0.22 & 8.91 &
    0.08 & 3.96 \\
    DRDG R. & 0.31 & 5.01 &
    0.15 & 3.06 &
    0.01 & 11.14 \\
    \hline
    \textbf{Category} & \multicolumn{2}{c}{\textbf{Llama3-8B-it - noise 0.1}} & \multicolumn{2}{c}{\textbf{Llama3-8B-it - noise 0.2}} & \multicolumn{2}{c}{\textbf{Llama3-8B-it - noise 0.3}} \\
    \hline
    & F1 (avg) & Time (s) & F1 (avg) & Time (s) & F (avg) & Time (s) \\
    \hline
    MIXED & 0.00 & 62.54 &
    0.00 & 176.31 &
    0.02 & 51.55 \\
    CHAIN & 0.00 & 38.86 &
    0.21 & 12.84 &
    0.06 & 29.31 \\
    CHAIN R. & 0.00 & 31.39 &
    0.32 & 34.90 &
    0.08 & 57.75 \\
    RDG & 0.00 & 41.42 &
    0.00 & 18.42 &
    0.07 & 55.80 \\
    RDG R. & 0.21 & 20.86 &
    0.20 & 36.45 &
    0.00 & 40.70 \\
    DRDG & 0.00 & 76.70 &
    0.08 & 45.48 &
    0.04 & 60.10 \\
    DRDG R. & 0.00 & 25.96 &
    0.00 & 43.88 &
    0.00 & 14.61 \\
    \hline
    \textbf{Category} & \multicolumn{2}{c}{\textbf{Mixtral-8x7B-It-v0.1 - noise 0.1}} & \multicolumn{2}{c}{\textbf{Mixtral-8x7B-It-v0.1 - noise 0.2}} & \multicolumn{2}{c}{\textbf{Mixtral-8x7B-It-v0.1 - noise 0.3}} \\
    \hline
    & F1 (avg) & Time (s) & F1 (avg) & Time (s) & F1 (avg) & Time (s) \\
    \hline
    MIXED & 0.36 & 34.04 &
    0.21 & 65.83 &
    0.20 & 71.60 \\
    CHAIN & 0.49 & 50.07 &
    0.00 & 45.15 &
    0.16 & 45.02 \\
    CHAIN R. & 0.47 & 29.56 &
    0.00 & 87.16 &
    0.08 & 69.38 \\
    RDG & 0.60 & 75.42 &
    0.05 & 35.68 &
    0.00 & 101.54 \\
    RDG R. & 0.48 & 74.58 &
    0.00 & 123.80 &
    0.00 & 69.17 \\
    DRDG & 0.10 & 90.51 &
    0.28 & 130.16 &
    0.12 & 82.51 \\
    DRDG R. & 0.15 & 60.61 &
    0.00 & 97.11 &
    0.00 & 64.92 \\
    \hline
    \end{tabular}
    \caption{Performance metrics for various categories under different noise conditions.}
    \label{tab:llm_res}
    \end{table*}

\begin{figure}[htb!]
        \centering
        \includegraphics[width=1.05\linewidth]{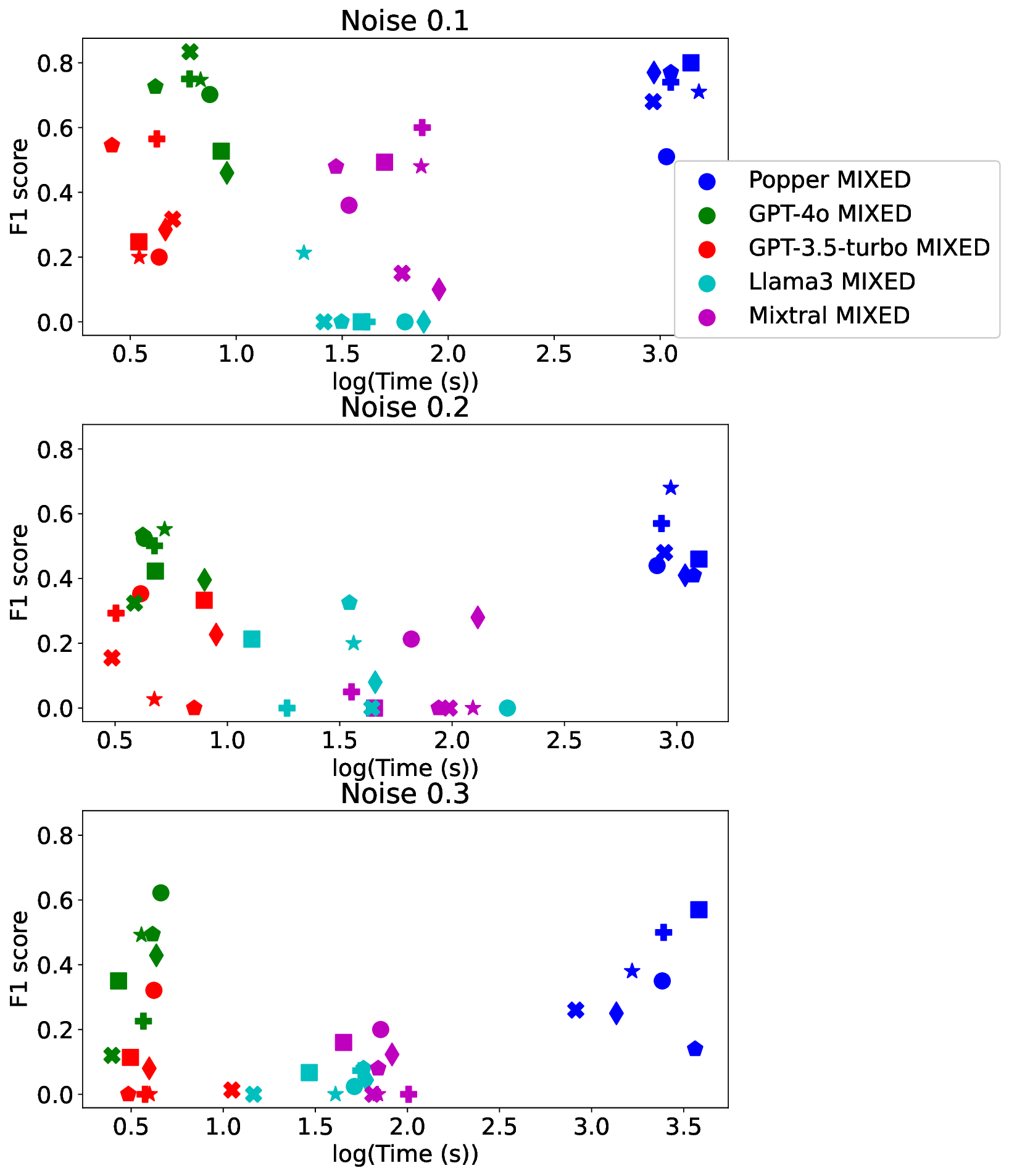}
        \caption{Relationship between the F1 score and the logarithm of processing time (in seconds) for five different mixed models—Popper, GPT-4, GPT-3.5-turbo, Llama3, and Mixtral—across three noise levels: 0.1, 0.2, and 0.3 and each rule set category: CHAIN ($\bullet$), CHAIN R.($\square$), RDG($\pentago$), RDG R.(\textbf{+}), DRDG($\bigstar$), DRDG R.($\diamondsuit$), and MIXED(\textbf{X}). Each subplot corresponds to a different noise level, showing how each model's performance and processing time vary with increasing noise. While Popper always takes more time to develop a theory, the other two levels (0.5 - 1.0, 1.0 - 2.5) correspond to different execution environments. Time variance changes in opposite ways w.r.t. noise on Popper vs. the LLMs.}
        \label{fig:f1_time}
\end{figure}

GPT-4o shows stable performance with moderate to high accuracy but is sensitive to noise, especially in RDG and DRDG. For instance, its F1 score in RDG drops from 0.75 at noise level 0.1 to 0.25 at noise level 0.3. GPT-3.5-Turbo does not perform well with complex categories like RDG and DRDG under noise, with an F1 score of 0 at noise level 0.3 in RDG.

Mixtral-8x7B-Instruct-v0.1 shows high variability w.r.t. noise, performing reasonably well in RDG (0.64 F1 at noise level 0.1, dropping to 0.43 at noise level 0.3) but with significant time consumption, especially in DRDG (130.160 seconds at noise level 0.2). It does not perform well with complex rule sets like DRDG across all noise levels.

Llama3-8B instruct has low accuracy across most categories, with slight improvement at higher noise levels but increased time consumption. At noise level 0.1, it achieves an F1 score above 0 only in RDG R. It often fails to produce valid theories or introduces new predicates incorrectly. For instance, the rule \textit{p(X, Y) - p2(X, Y); p0(X, Y); p4(X, Y); p9(X, Y).} is valid, but the predicate $p$, in the head of the rule, does not exist in the BK, neither is the target predicate. Llama3-8B was the only model to exhibit this pattern.


The models generally present higher initial accuracy on recursive ($R$) categories, but are more sensitive to noise on them, leading to larger performance drops. For example, on DRDG-R, GPT-4o's F1 score drops from 0.83 at noise level 0.1 to 0.120 at noise level 0.3. Non-recursive categories like CHAIN present more stable performance.



\paragraph{Time variance has an inverse relation w.r.t. noise levels on LLMs when compared with Popper.}
As the noise increases, Popper may take less time to induce a theory based on the remaining relevant data, as indicated by the scattering pattern progression in Figure~\ref{fig:f1_time}, while the LLMs are more likely to take longer to process it. Detailed values are presented in Tables~\ref{tab:popper_res} and \ref{tab:llm_res}.

\section{Prompt templates} \label{sec:prompts}

For the iterative LM theory refinement method, the following template was used for the initial prompt:

\paragraph{Prompt $p_1$:} \label{pt1}

\begin{spverbatim}
Induce a theory based on background knowledge, positive and negative examples. Write it in Prolog. Do not give an explanation. Answer only the theory.

BK:
{BK}

Examples:
{positive and negative examples}
\end{spverbatim}

~\\

For the refinement steps, the following prompt template was used:

\paragraph{Prompt $p_2...p_n$:} \label{pt2}
\begin{spverbatim}
The theory scored:
accuracy = {acc} precision = {precision} recall = {recall} f1 = {f1}

and got these examples wrongly:
{examples that were misclassified}

Refine the theory. Answer only the theory.

\end{spverbatim}

The prompt templates were designed to be objective and minimal, containing only the necessary instructions and data for solving the task. They were adjusted using a small sample of inputs, to minimise the syntax errors across all models. The same prompt templates were used across all language models.

\section{Reproducibility}

Upon acceptance of this paper, we will release all code and data associated with our study. The raw data, along with detailed instructions for data processing, are accessible via the provided repository link. Any proprietary tools or materials used in this study are either commercially available or provided under a reasonable request. By ensuring that all aspects of this research are openly accessible, we invite the scientific community to replicate our findings and build upon this work, fostering a collaborative and transparent scientific environment.




\section{Dataset Generation Process}
\label{appendix:dataset_generation}

Below, we provide a detailed description of each of the parameters used for the dataset generation process, along with their specific configurations.

\subsection*{Parameters}

\begin{itemize}
    \item \textbf{mindags}:
    \begin{itemize}
        \item \textbf{Definition}: Minimum number of generated DAGs. This parameter ensures that at least the specified number of DAGs is generated in the dataset.
        \item \textbf{Constraint}: Must be greater than 0.
    \end{itemize}

    \item \textbf{maxdags}:
    \begin{itemize}
        \item \textbf{Definition}: Maximum number of generated DAGs. This parameter sets an upper limit on the number of DAGs to be included in the dataset.
        \item \textbf{Constraint}: Must be greater than or equal to \texttt{mindags}.
    \end{itemize}

    \item \textbf{noise}:
    \begin{itemize}
        \item \textbf{Definition}: Represents the percentage of noise in the datasets. Noise here refers to the random perturbations added to the data.
        \item \textbf{Constraint}: Must be a value in the range [0, 1].
    \end{itemize}

    \item \textbf{owa (Open World)}:
    \begin{itemize}
        \item \textbf{Definition}: 
        The open-world degree indicates how many of the consequences of an initial set of relevant facts, called support facts, are missing from the data set. In other workds it indicates the percentage of consequences missing in the dataset. This parameter simulates incomplete data scenarios by randomly omitting a portion of the data.
        \item \textbf{Constraint}: Must be a value in the range [0, 1].
    \end{itemize}

    \item \textbf{missing}:
    \begin{itemize}
        \item \textbf{Definition}: Specifies the percentage of missing data in the dataset.
        \item \textbf{Constraint}: Must be a value in the range [0, 1].
    \end{itemize}

    \item \textbf{category}:
    \begin{itemize}
        \item \textbf{Definition}: Determines the type of the rule to be generated. The categories include different structural patterns and combinations.
        \item \textbf{Values}:
        \begin{itemize}
            \item Chain
            \item Rooted Directed Graph (DG)
            \item Disjunctive Rooted DG
            \item Mixed
            \item All of them with recursion
        \end{itemize}
    \end{itemize}
\end{itemize}

\section*{Illustrative Example: Family Knowledge Base}

To illustrate the dataset's construction, consider a simple example representing a small knowledge base about familial relationships. Here, the fact \textbf{parent(john, mary).} denotes that \textbf{John} is the parent of \textbf{Mary}. A corresponding rule might be expressed as: $\forall X, Y \ (\text{parent}(X, Y) \rightarrow \text{ancestor}(X, Y))$.

The initial dataset could be represented as follows:

\begin{minted}{prolog}
% Facts
parent(john, mary).
parent(mary, susan).
parent(john, michael).
parent(michael, robert).

% Rules
ancestor(X,Y) :- parent(X,Y).
ancestor(X,Y) :- parent(X,Z),ancestor(Z,Y).
\end{minted}

\subsection*{Scenario 1: Missing Data}

Assume that 20\% of the data is missing. The dataset would then be:

\begin{minted}{prolog}
% Facts with 20% missing
parent(john, mary).
% parent(mary, susan). % This fact is missing
parent(john, michael).
parent(michael, robert).

% Rules
ancestor(X,Y) :- parent(X,Y).
ancestor(X,Y) :- parent(X,Z),ancestor(Z,Y).
\end{minted}

\subsection*{Scenario 2: Noisy Data}

Alternatively, if 20\% of the data contains noise, the dataset might appear as follows:

\begin{minted}{prolog}
% Facts with 20% noise
parent(john, mary).
parent(mary, susan).
parent(john, michael).
parent(michael, robert).
%Below is a noisy fact
parent(michael, alice). 

% Rules
ancestor(X,Y) :- parent(X,Y).
ancestor(X,Y) :- parent(X,Z),ancestor(Z,Y).
\end{minted}

This methodical approach to dataset generation allows us to simulate a wide range of real-world conditions, providing a robust foundation for analyzing the effects of noise, missing data, and structural variations on the performance of our experiments.

\section{LLMs}

Table \ref{table:models} provides a summary of the main information about the models used in this study.

{
\small
\begin{table*}[ht!]
\centering
\begin{tabular}{lccccc}
\hline
\textbf{Model} & \textbf{Maintainer} & \textbf{Parameters} & \textbf{Hidden dim.} & \textbf{\# hidden layers} & \textbf{Context size} \\
\hline
LLaMA3-8B Instr.      & Meta-Llama   & 8B             & 4096 & 32 & 8K \\
Mixtral-8x7B Instr.   & Mistral AI   & 46.7B (sparse) & 4096 & 32 & 32K \\
Gemma-7B-IT             & Google       & 7B             & 3072 & 28 & 8K \\
GPT-3.5 Turbo           & OpenAI       & --            & -- & -- & 16K \\
GPT-4o                  & OpenAI       & --            & -- & -- & 128K \\
\hline
\end{tabular}
\caption{Main Information about the models evaluated in this study. All models tested are auto-regressive decoder-only, with Mixtral-8x7B Instruct being a Sparse Mixture of Experts (SMoE). The original, non-quantised versions were used.}
\label{table:models}
\end{table*}
}

\section{Models Output} \label{sec:model_outputs}

The GPT-4o and GPT-3.5-turbo models have been demonstrated to consistently generate valid theories, thereby ensuring the successful execution of the Prolog code they produce. To illustrate, the following displays a theory induced by GPT-4o.

\begin{minted}{prolog}
p10(A,B):-p8(A,B).
p10(A,B):-p1(A,B).
p10(A,B):-p7(A,B).
\end{minted}

Nevertheless, a recurrent pattern has been identified in the theories generated by GPT, namely the rewriting of rules in which the variable is interchanged. To illustrate, the following example is provided.

\begin{minted}{prolog}
p1(A, B) :- p2(A, B).
p1(A, B) :- p2(B, A).
\end{minted}

Furthermore, GPTs, particularly GPT-4o, are highly effective at identifying the relevant predicate for a given theory, disregarding the noise, the irrelevant facts added on purpose in the dataset. The following is an exemplar rule:

\begin{minted}{prolog}
p1(X0,X1) :- p2(X0,X1).
p2(X0,X1) :- p4(X1,X2),p0(X0,X2).
\end{minted}

The initial prompt identifies the predicate p2

\begin{minted}{prolog}
p1(A, B) :- p2(A, B).
\end{minted}

The refinement identifies the predicate p0

\begin{minted}{prolog}
p1(A, B) :- p2(A, B).
p1(A, B) :- p0(A, B).
\end{minted}

Subsequently, the predicate p4 is identified.

\begin{minted}{prolog}
p1(A, B) :- p2(A, B).
p1(A, B) :- p0(A, B).
p1(A, B) :- p4(A, B).
\end{minted}

However, the same degree of precision could not be obtained from Llama3-8B, which may not consistently generate Prolog code that adheres to the necessary syntactical or logical constraints, potentially leading to errors during execution. To illustrate, consider the following Prolog theory:

\begin{minted}{prolog}
theory :-
    p(X, Y) :- p1(X, Y); p3(X, Y); 
    p4(X, Y); p7(X, Y); p8(X, Y); 
    p0(X, Y),
    not neg(p(X, Y)),
    (pos(p(X, Y)) - true; fail).
\end{minted}

The models in question autonomously created the head of the clause "theory," as well as the predicates "p" and "pos," which should not exist. 

Additionally, Mixtral demonstrated satisfactory performance, although it exhibited a proclivity to insert the theory at the outset of the output. Although the output was generally valid, the quality of the generated theories was not as robust as that of GPT-4o, particularly in more intricate recursive scenarios such as RDG and DRDG. Additionally, it was also able to identify the relevant predicates, but their arrangement was not optimal. For example, in the same example as GPT-4o, the correct predicates were identified, but their arrangement was not optimal.

\begin{minted}{prolog}
p1(X,Y) :-
 p0(X,Y),
 \+ p2(X,Y),
 \+ p4(X,Y).
\end{minted}

Furthermore, this model produces an excessive number of rules, particularly in more intricate rule sets such as RDG, DRDG, and MIXED, both recursive and non-recursive. It also introduces pos or neg predicates that are erroneous and should not exist, in 20\% of the results in these categories. The following example demonstrates a theory generated by Mixtral-8x7B with these issues: 

\begin{minted}{prolog}
theory :-
dif(X, Y),
p3(X, Y), p5(Y, Y), p7(Y, _),
\+ pos(p2(X, Y)),
\+ neg(p2(X, Y)),
asserta(pos(p2(X, Y))).

theory :-
dif(X, Y),
p4(X, Y), p5(Y, Y), p7(Y, _),
\+ pos(p2(X, Y)),
\+ neg(p2(X, Y)),
asserta(pos(p2(X, Y))).

theory :-
dif(X, Y),
p6(X, Y), p5(Y, Y), p7(Y, _),
\+ pos(p2(X, Y)),
\+ neg(p2(X, Y)),
asserta(pos(p2(X, Y))).

theory :-
dif(X, Y),
p10(X, Y), p5(Y, Y), p7(Y, _),
\+ pos(p2(X, Y)),
\+ neg(p2(X, Y)),
asserta(pos(p2(X, Y))).

theory :-
X \= Y,
p5(X, X),
\+ pos(p2(X, Y)),
\+ neg(p2(X, Y)),
asserta(pos(p2(X, X))).
\end{minted}

Finally, Gemma 7B did not produce a valid theory. While Llama-3-8B does not entirely conform to the characteristics of a theory, it nonetheless approximates a theory to a certain degree. In comparison, Gemma 7B's output lacked the elements necessary to be considered even a preliminary valid theory. The following is an example of a Gemma's output.

\begin{spverbatim}
**Theory:**

The facts in the knowledge base indicate that the predicate p9(cX, cY) is true for the following pairs of facts:

- c41 and c17
- c13 and c52
- c54 and c7
- c62 and c61
- c71 and c75
- c24 and c48
- c79 and c67
- c50 and c46
- c70 and c60
- c52 and c51
- c81 and c71
- c2 and c14
- c30 and c44
- c78 and c72
- c81 and c35

However, the predicate p9(c55, c48) is false.
\end{spverbatim}

\end{document}